\documentclass[final,5p,times,twocolumn]{elsarticle}

\usepackage{amssymb}

\usepackage{graphicx}
\usepackage{amsmath}
\usepackage{amssymb}
\usepackage{booktabs}
\usepackage{multirow}
\usepackage[table]{xcolor}
\usepackage{algorithm}
\usepackage{algorithmic}
\usepackage{multirow}
\usepackage[table]{xcolor}
\usepackage{newfloat}
\usepackage{listings}
\usepackage[numbers]{natbib}
\usepackage{graphicx}
\usepackage{rotating}
\usepackage{makecell}
\usepackage{colortbl} 
\usepackage{wrapfig}
\usepackage{flushend}
\usepackage{url}
\definecolor{lgray}{RGB}{230,230,230}

\usepackage{titlesec}

\titlespacing\section{0pt}{12pt plus 4pt minus 2pt}{0pt plus 2pt minus 2pt}
\titlespacing\subsection{0pt}{12pt plus 4pt minus 2pt}{0pt plus 2pt minus 2pt}
\titlespacing\subsubsection{0pt}{12pt plus 4pt minus 2pt}{0pt plus 2pt minus 2pt}

\journal{Nuclear Physics B}

\begin{document}

\begin{frontmatter}

    
    
\title{NIV-SSD: Neighbor IoU-Voting Single-Stage Object Detector From Point Cloud}
    

\author[1]{Shuai~Liu} 
\author[2]{Di~Wang\corref{cor1}}
\author[2]{Quan~Wang}
\author[1]{Kai~Huang}

\address[1]{School of Computer Science and Engineering, Sun Yat-sen University, Guangzhou 510006, China}
\address[2]{School of Computer Science and Technology, Xidian University, Xi'an 710071, China}
\cortext[cor1]{Corresponding Author: (wangdi@xidian.edu).}
    
\begin{abstract}
Previous single-stage detectors typically suffer the misalignment between localization accuracy and classification confidence. To solve the misalignment problem, we introduce a novel rectification method named \textbf{\textit{neighbor IoU-voting} (\textit{NIV})} strategy. Typically, classification and regression are treated as separate branches, making it challenging to establish a connection between them. Consequently, the classification confidence cannot accurately reflect the regression quality. NIV strategy can serve as a bridge between classification and regression branches by calculating two types of statistical data from the regression output to correct the classification confidence. Furthermore, to alleviate the imbalance of detection accuracy for complete objects with dense points (easy objects) and incomplete objects with sparse points (difficult objects), we propose a new data augmentation scheme named \textbf{\textit{object resampling}}. It undersamples easy objects and oversamples difficult objects by randomly transforming part of easy objects into difficult objects. Finally, combining the \textit{NIV} strategy and object resampling augmentation, we design an efficient single-stage detector termed \textbf{NIV-SSD}. Extensive experiments on several datasets indicate the effectiveness of the \textit{NIV} strategy and the competitive performance of the NIV-SSD detector. The code will be available at \url{https://github.com/Say2L/NIV-SSD}.
\end{abstract}

\begin{keyword}
3D object detection \sep point cloud \sep single-stage detection
    
    
\end{keyword}
    
\end{frontmatter}


\section{Introduction}
LiDAR plays an important role in the perception system of autonomous driving. Compared to camera images, 3D point clouds from LiDAR can provide precise depth information and robust environment information under different levels of light. Hence, LiDAR-based 3D object detection has attracted much attention in recent years. 

Misalignment between classification confidence and localization accuracy frequently poses a challenge for 3D object detectors \cite{alca,cia-ssd}. For instance, a predicted bounding box of high quality may exhibit low classification confidence, whereas a poor-quality bounding box may have high classification confidence. This discrepancy can lead to the filtering out of high-quality bounding boxes during the non-maximum suppression (NMS) process, while retaining low-quality ones, thereby degrading the overall detection accuracy.

Typically, two-stage detectors \cite{voxel-rcnn, pv-rcnn} are less affected by the misalignment problem compared to single-stage detectors \cite{pointpillars, second}. Because two-stage detectors rely on region proposals generated by the first-stage network to predict the Intersection over Union (IoU) between predicted bounding boxes and ground truth boxes as the final confidence in the second stage. Though the predicted IoU is closer to the localization accuracy compared to the classification confidence, the computational cost is greatly raised due to the introduction of the second-stage network.

To solve the problem of misalignment, the single-stage detector SA-SSD \cite{sa-ssd} divides a predicted bounding box into grids, then uses an interpolation method to obtain the confidence for each grid point on classification maps, finally obtains the confidence of the bounding box by averaging the confidences of all grid points. However, the interpolation approach of SA-SSD is very complex. CIA-SSD \cite{cia-ssd} appends an IoU prediction branch to a single-stage network. It utilizes IoU predictions to help correct classification confidences. Nevertheless, single-stage detectors cannot extract features from region proposals, so the predicted IoUs are not as accurate as those of two-stage detectors. 

\begin{figure}[t]
    \centering
    \includegraphics[width=0.9\columnwidth]{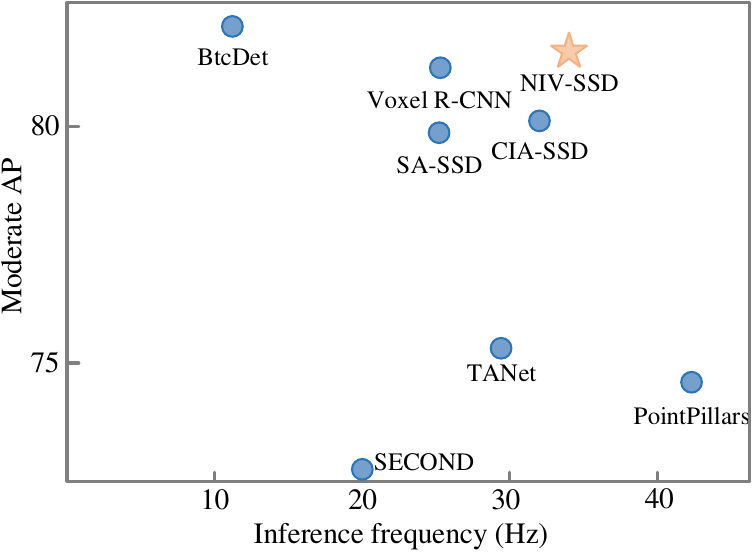}
    \caption{Comparisons on speed and accuracy. Results are obtained on 3D car detection in the KITTI \textit{test} set.}
    \label{fig1}
\end{figure}

\begin{figure}[t]
    \centering
    \includegraphics[width=1.0\columnwidth]{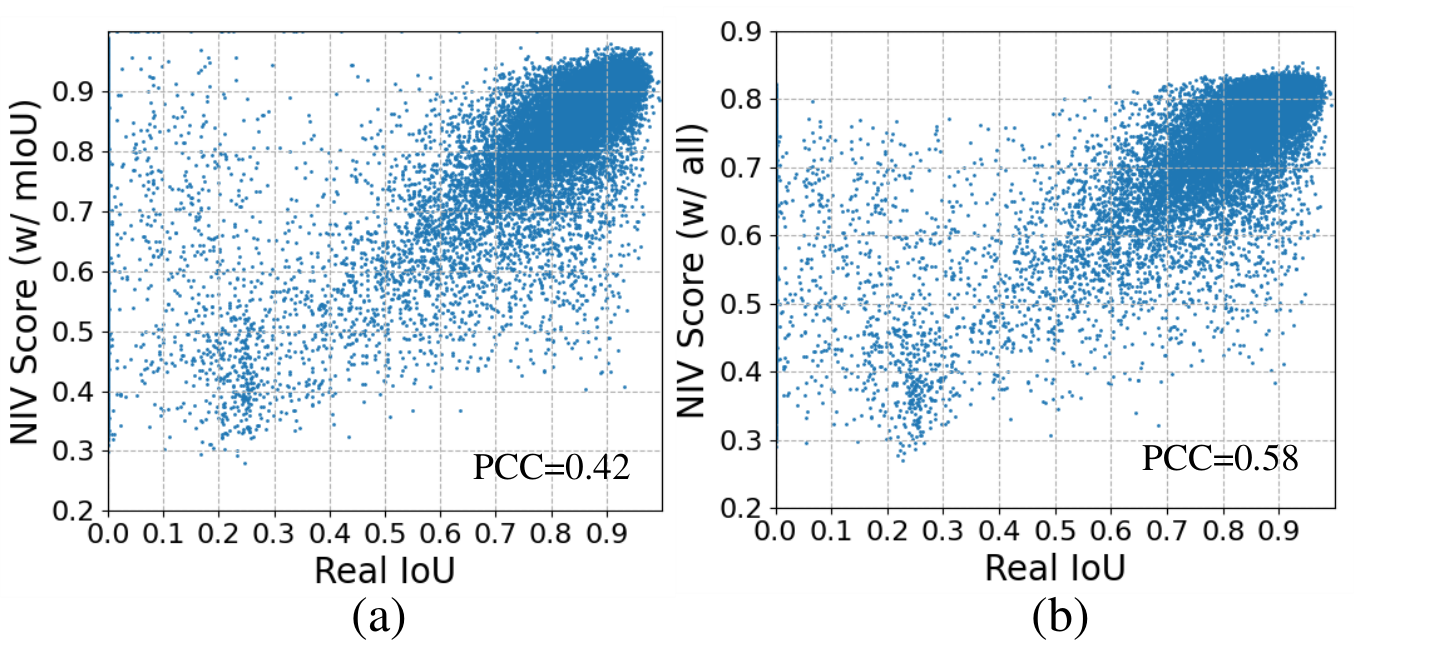}
    \caption{Scatterplots: (a) real IoU vs. NIV score (w/ mIoU) which denotes the mean IoU between a predicted box and its neighbors; and (b) real IoU vs. NIV score (w/ all) which denotes the combination of the mean IoU and the number of neighbors. ``PCC" denotes the Pearson correlation coefficient. }
    \label{fig2}
\end{figure}

To further tackle the misalignment problem in single-stage detectors, we propose an elegant post-processing confidence rectification method named \textit{neighbor IoU-voting (NIV)} strategy, which requires no modification to the network structure and only incurs minimal computational overhead. Our key idea is based on the following findings: (i) objects in point clouds do not overlap with each other, thus predicted neighbor bounding boxes\footnote{If the IoU between two bounding boxes is higher than a threshold, the two bounding boxes are considered to be neighbors. } are generally related to one ground-truth object; (ii) a predicted bounding box with higher localization quality generally has more overlapped neighbor bounding boxes and a larger mean IoU with its neighbor bounding boxes. Thus, we can rectify the confidence of a bounding box by referring to the number of its neighbor bounding boxes and the mean IoU between it and its neighbors. As presented in Figure~\ref{fig2}, the mean IoU between predicted boxes and their neighbors is positively correlated to the real IoU between predicted boxes and ground truth boxes. Additionally, the Pearson correlation coefficient (PCC) between real IoUs and NIV scores is higher when the number of neighbors is considered. The above demonstrates that both the mean IoU and the number of neighbors are useful statistical data, while they have not been considered in prior works.

Furthermore, we propose a new data augmentation scheme named object resampling which randomly transforms objects with dense points and minor occlusion (easy objects) into objects with sparse points and severe occlusion (difficult objects). The motivation behind this augmentation scheme is the finding that detectors are generally more sensitive to easy objects and biased against difficult objects. Hence, we increase the number and diversity of difficult objects through the object resampling augmentation to improve the detection accuracy for difficult objects. Combining the \textit{NIV} strategy and object resampling augmentation, we design a single-stage detector named NIV-SSD. As demonstrated in Figure~\ref{fig1}, our NIV-SSD detector strikes a harmonious balance between speed and accuracy.

The contributions of this work can be summarized as follows:
\begin{itemize}
    \setlength{\itemsep}{0pt}
    \setlength{\parsep}{0pt}
    \setlength{\parskip}{0pt}
    \item An elegant post-processing rectification strategy named NIV is proposed to align the classification confidence with the localization quality of predicted bounding boxes. 
    \item A new data augmentation scheme named object resampling is introduced to improve the detection accuracy of detectors for difficult objects. 
    \item An efficient single-stage detector named NIV-SSD is proposed. Extensive experiments on several datasets demonstrate the effectiveness and generality of the \textit{NIV} strategy and a good balance between the speed and accuracy of our NIV-SSD detector.
\end{itemize}

\section{Related Work}
The LiDAR-based 3D object detectors can be divided into two categories: two-stage detectors and single-stage detectors. Two-stage detectors have an additional refinement stage for rectifying predicted bounding boxes and classification confidences utilizing region-proposal-aligned features. Therefore, two-stage detectors typically achieve better detection accuracy compared to single-stage detectors. However, due to the extra refinement network of two-stage detectors, they tend to have a high latency, which is unacceptable for autonomous driving systems with real-time requirements. Single-stage detectors usually have faster inference speed but are inferior to two-stage detectors in terms of detection accuracy.

\subsection{Two-Stage Detectors}
PointRCNN \cite{pointrcnn} utilizes PointNet++ \cite{pointnet++} to produce proposals from raw points, then refines bounding boxes in the second stage. Part-$\mathit{A}^{2}$ \cite{parta2} exploits 3D intra-object part locations to aid the second-stage refinement. Fast Point R-CNN \cite{fastpr} utilizes a voxel-based network to obtain initial predictions. Then it refines predictions by coordinates and semantic features of internal points of proposals. PV-RCNN \cite{pv-rcnn} which is similar to Fast Point R-CNN uses farthest point sampling (FPS) to sample a small number of key points in the second stage to reduce latency. Voxel R-CNN \cite{voxel-rcnn} exploits 3D voxel features in the 3D backbone to replace features of raw points for the second-stage refinement. CenterPoint \cite{centerpoint} refines proposals using point features around the center of predicted bounding boxes. BtcDet \cite{btcdet} utilizes an extra network to predict the probability of occupancy that indicates if a region contains an object, and then combines the probability map to generate initial predictions and refine bounding boxes.

\begin{figure*}[t]
    \centering
    \includegraphics[width=0.95\textwidth]{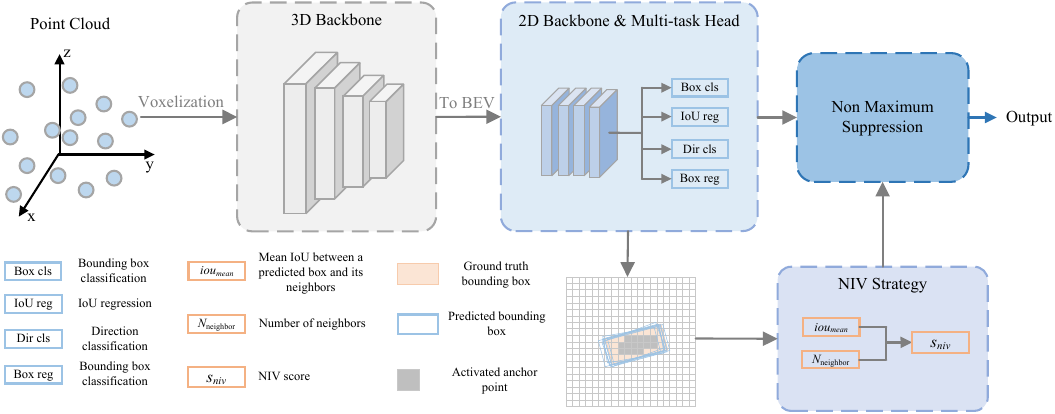}
    \caption{The detection pipeline of our NIV-SSD. First, a point cloud is transformed into voxels. Next, the voxels are fed to a 3D backbone which is composed of 3D sparse convolutions. A 2D feature map is generated by the 3D backbone. Then, a 2D backbone is used to extract features from the 2D feature map, and a multi-task head module is utilized to produce multi-task predictions. Finally, the neighbor IoU-voting (NIV) strategy is adopted to rectify classification confidences, and NMS is used to filter redundant predictions. }
    \label{fig3}
\end{figure*}

\subsection{Single-Stage Detectors} 
VoxelNet \cite{voxelnet} encodes voxel features by PointNet \cite{pointnet} and then extracts features from 3D feature maps by 3D convolutions. SECOND \cite{second} proposes 3D sparse convolution to efficiently encode sparse voxel features. PointPillars \cite{pointpillars} divides a point cloud into pillar voxels to avoid using 3D convolution layers, thus achieving high inference speed. 3DSSD \cite{3dssd} discards upsampling layers and the refinement network commonly used in point-based methods, thus significantly improving the inference speed. IA-SSD \cite{ia-ssd} gradually removes background points during undersampling and preserves foreground points that provide important information, so as to effectively reduce the size of point clouds without loss of precision. SE-SSD \cite{se-ssd} uses a teacher model to provide soft labels to assist in supervising the training of a student model.

\subsection{Solutions for Misalignment}
Compared to two-stage detectors, single-stage detectors generally suffer from a worse misalignment problem. To solve the problem, SA-SSD \cite{sa-ssd} proposes a part-sensitive warping operation that divides a predicted bounding box into grids and obtains the final confidence by averaging the confidences of several grid points. And CIA-SSD \cite{cia-ssd} exploits an extra IoU prediction for confidence rectification. Similar to CIA-SSD, some approaches like Fitness NMS \cite{fitnessNMS}, IoU-Net \cite{iounet}, MS R-CNN \cite{masksrcnn}, FCOS \cite{fcos} and IoU-aware \cite{iou-aware} utilize a separate branch to perform localization quality estimation in the form of IoU or centerness score. GFL \cite{gfl} proposes an improved focal loss named quality focal loss (QFL) which uses consistent IoU values as labels. Therefore QFL can obtain classification-IoU joint representations for directly presenting the quality of predicted bounding boxes. Though these methods rectify the classification confidence to some extent, the misalignment problem is still severe. In this paper, we propose a single-stage detector NIV-SSD which introduces an elegant strategy to further address the misalignment problem. Details about NIV-SSD are described in the methodology section.

\subsection{Data Augmentation for Point Cloud}
Traditional data augmentation methods for point clouds include translation, rotation, flipping, and scaling. Recently, several other data augmentation methods have been proposed. SECOND \cite{second} suggests creating a database of object points, from which objects are randomly selected during training and then added to the current point cloud scene. Generally, there are significantly more vehicle objects than objects from other categories, resulting in a long-tailed distribution of object categories. To address this issue, CBGS \cite{cbgs} proposes a class-balanced grouping and sampling strategy to ensure balanced objects for each category. Furthermore, SE-SSD \cite{se-ssd} introduces a share-aware data augmentation scheme to enhance object diversity. Unlike previous data augmentation schemes, our object resampling scheme focuses on the balance between easy and difficult objects. More details about our method will be provided in the next section. 

\section{Neighbor IoU-Voting Single-Stage Detector}

\subsection{Task Setup}
Given a LiDAR point cloud $\{p_1, p_2, ..., p_n\}$, the purpose of LiDAR-based 3D object detection is to detect objects such as vehicles, non-motorized vehicles, and pedestrians in the point cloud. 
Let $(x, y, z)$ and $i$ denote the coordinates and reflection intensity of a point, respectively. NIV-SSD first voxelizes the point cloud and then calculates the mean coordinates and intensities of points in each voxel. 
Let $(\bar{x}, \bar{y}, \bar{z}, \bar{i})$ denote the initial feature of a voxel. 
\subsection{Overall Framework}

The overview of our NIV-SSD pipeline is shown in Figure~\ref{fig3}. The network of NIV-SSD is composed of three parts including a 3D backbone, a 2D backbone, and a multi-task head.

\noindent \textbf{3D Backbone Network.}
The 3D backbone is used to extract features from sparse voxels and convert 3D feature volumes into bird's eye view (BEV) representations. Unlike most previous approaches \cite{second, cia-ssd}, the 3D backbone of NIV-SSD contains residual connections. It is composed of four blocks, each containing one sparse convolution (SC) or one submanifold sparse convolution and several residual submanifold sparse convolutions (RSSC). The RSSC consists of two submanifold sparse convolutions and a residual connection. Though the residual connection enhances the feature extraction capability of the model, it introduces additional latency. To balance the accuracy and speed of the model, we rescaled the width and depth of the 3D backbone used in SECOND \cite{second}. We call the modified 3D backbone lite 3DSparseResNet. Specifically, the channels and numbers of RSSC in four blocks are \{16, 32, 64, 64\} and \{1, 1, 2, 2\}, respectively. Only the first block does not contain sparse convolution, which consists of a submanifold sparse convolution followed by an RSSC, and other blocks are composed of a sparse convolution followed by several RSSCs. Finally, the 3D voxel features are concatenated along the height dimension to form a BEV feature map.

\begin{figure}[t]
    \centering
    \includegraphics[width=0.8\columnwidth]{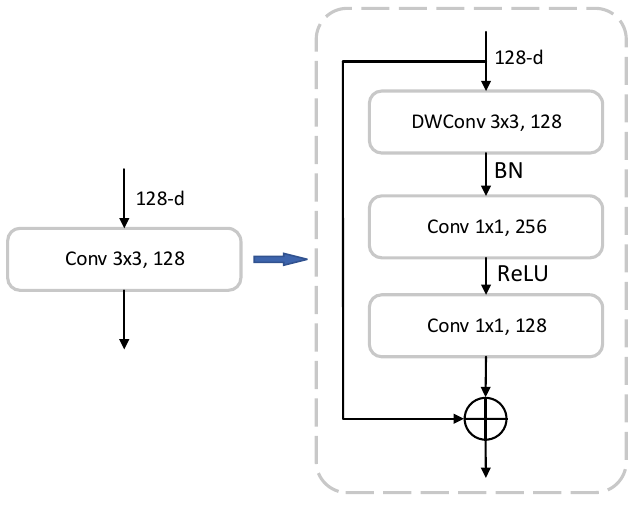}
    \caption{A diagram of replacing a traditional convolution layer with a ConvNeXt block.}
    \label{fig4}
\end{figure}

\noindent\textbf{2D Backbone Network.}
The design of the 2D backbone network bears resemblance to previous works such as \cite{second, voxel-rcnn, sa-ssd}. The 2D backbone network consists of two stages. The first stage focuses on extracting low-level spatial features, where the input and output feature maps have the same resolution. The second stage is dedicated to extracting high-level semantic features. In \cite{second, voxel-rcnn, sa-ssd}, the 2D backbone employs standard $3 \times 3$ convolution layers. However, our NIV-SSD replaces the $3 \times 3$ convolution layers with modified ConvNeXt blocks \cite{convnext} that are tailored to adapt to the 3D object detection task. As shown in Figure~\ref{fig4}, the ConvNeXt block comprises a depthwise convolution layer, followed by two pointwise convolution layers. A batchnorm layer is appended to the depthwise convolution layer, and a Rectified Linear Unit (ReLU) is applied to the first pointwise convolution layer. Additionally, a shortcut connection exists between the input and output. Specifically, the two stages employ ConvNeXt blocks with channel numbers of {128, 256} and {5, 5}, respectively. The first pointwise convolution layer expands the number of channels to twice the original size, while the second pointwise convolution layer reduces it back to the original size.

\noindent \textbf{Multi-Task Head.}
The misalignment between localization accuracy and classification confidence is a common issue encountered in single-stage detectors. To address this problem, we adopt an IoU prediction branch in the multi-task head, following the approach proposed in \cite{cia-ssd}. More specifically, the output feature map of the 2D backbone undergoes four $1 \times 1$ convolution layers in parallel, generating separate predictions for each task. The loss function employed in NIV-SSD is identical to that used in \cite{cia-ssd}.

\begin{algorithm}[t]
    \caption{Neighbor IoU-Voting Strategy}
    \label{Alg:alg1}
    \begin{algorithmic}[1]
    \REQUIRE ~~\\
    Predicted bounding boxes $\mathcal{B}_0$ of one category with the size of $N \times 7$, where $N$ is the number of bounding boxes, and $(x, y, z, w, l, h, r)$ is the parameters of a bounding box, $(x, y, z)$ denotes box center, $(w, l, h)$ denotes box size, and $r$ denotes orientation angle;\\
    Predicted classification confidence values $\mathcal{C}$ of the corresponding predicted bounding boxes with the size of $N \times 1$;\\
    BEV area of anchor $area_{bev}$;\\
    Final confidence score threshold $score\_thres$;\\
    IoU threshold $iou\_thres$;\\
    $\mathcal{B}_0 = \{\mathbf{b}_1, \mathbf{b}_2, \cdots , \mathbf{b}_n\}$; $\mathcal{C} =  \{c_1, c_2, \cdots , c_n\}$; \\

    \ENSURE ~~\\
    Selected bounding boxes $\mathcal{B} = \emptyset$;\\
    Rectified confidence values $\mathcal{S} = \emptyset$ of the corresponding selected bounding boxes;
    \FOR{$i = 0, 1, \cdots, N$}
        \STATE{$iou_{all} = 0, N_{neighbor} = 0$;}
        \FOR{$j = 0, 1, \cdots, N$}
            \IF{$\mathrm{IoU}(\mathbf{b}_i, \mathbf{b}_j) > iou\_thres$}
                \STATE{
                    $iou_{all} \gets  iou_{all} + \mathrm{IoU}(\mathbf{b}_i, \mathbf{b}_j)$;
                }
                \STATE{
                    $N_{neighbor} \gets  N_{neighbor} + 1$;
                }
            \ENDIF
        \ENDFOR
        \STATE{
            $iou_{mean} = \frac{iou_{all}}{N_{neighbor}};$
        }
        \STATE{
            $N_{neighbor} = N_{neighbor} \cdot \frac{area_{bev}}{\mathbf{b}_{i}[3]\cdot \mathbf{b}_{i}[4]};$
        }
        \STATE{
            $s_{niv} = \frac{N_{neighbor}}{N_{neighbor} + 1} \cdot iou_{mean};$
        }
        \STATE{
            $s = s_{niv} \cdot c_i;$
        }
        \IF{$s > score\_thres$}
            \STATE{
                $\mathcal{B} \gets \mathcal{B} \cup \mathbf{b}_i$;\\
                $\mathcal{S} \gets \mathcal{S} \cup s$;
            }
        \ENDIF
    \ENDFOR
    \RETURN $\mathcal{B}, \mathcal{S}$
\end{algorithmic}
\end{algorithm}

\subsection{Neighbor IoU-Voting Strategy}
The classification and regression branches play distinct roles in object classification and localization, respectively. These branches operate independently, leading to a discrepancy between classification confidence and localization accuracy. To tackle this issue, the IoU-aware method \cite{iou-aware,cia-ssd} introduces an additional IoU branch to the network, establishing a connection between the classification and regression branches. While the IoU prediction in single-stage detectors helps to rectify the classification confidence to some extent, it still falls short compared to two-stage detectors. This is because two-stage detectors can evaluate the IoUs between predicted bounding boxes and ground truth bounding boxes by utilizing region-proposal-aligned features. In contrast, single-stage detectors directly perform IoU regression on the output feature map.


To further enhance the confidence prediction of single-stage detectors, we introduce the neighbor IoU-voting (NIV) strategy. This strategy leverages two types of statistical data derived from the regression output to refine the classification confidence. The underlying idea behind the NIV strategy is that bounding boxes with higher localization accuracy tend to have more neighbor bounding boxes (abbreviated as neighbors in the following for simplicity) and exhibit greater overlap with their neighbors, leading to a larger mean IoU. This relationship is illustrated in Figure~\ref{fig2}. Furthermore, our observations indicate that objects in point clouds typically do not overlap with each other, meaning that neighbors are often associated with the same ground-truth object. Drawing from these insights, we propose the neighbor IoU-voting strategy, which takes into account the contribution of neighbors to rectify the classification confidence. 

\begin{figure}[!]
    \centering
    \resizebox{\linewidth}{!}{
    \includegraphics[width=0.9\columnwidth]{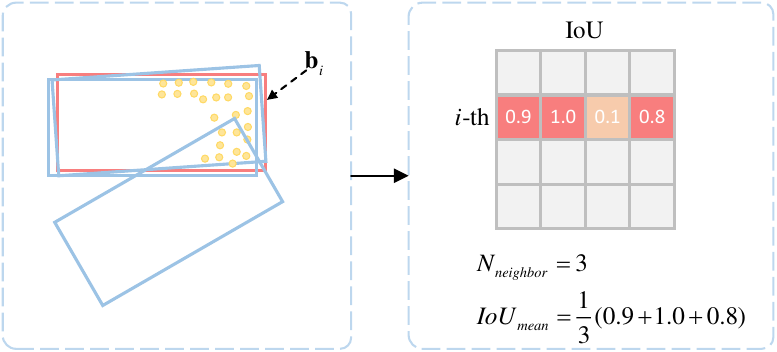}}
    \caption{A simple example of \textit{NIV} calculating. }
    \label{fig_niv_example}
\end{figure}

The procedure of our neighbor IoU-voting strategy is outlined in Algorithm~\ref{Alg:alg1}. These steps can be summarized as follows: first, calculate the IoU between each pair of all predicted bounding boxes; then, for each bounding box, count the number of its neighbors and calculate the mean IoU between it and its neighbors; next, rectify the classification confidence utilizing the number of neighbor bounding boxes and the mean IoU value as the step 10 and 11 in Algorithm~\ref{Alg:alg1}; finally, filter out bounding boxes with low rectified confidence values. Figure~\ref{fig_niv_example} shows a simple case of how to obtain the two statistical data of \textit{NIV}. Without loss of generality, the classification confidence in Algorithm~\ref{Alg:alg1} can be replaced by the confidence from other rectified methods such as IoU-aware \cite{cia-ssd}. 

\subsection{Object Resampling Data Augmentation}


Due to the rotational scanning nature of LiDAR, the density of points in point clouds varies depending on the distance. Areas closer to the LiDAR exhibit higher point density, while areas farther away have sparser points. Moreover, objects in point clouds often encounter varying degrees of occlusion, stemming from external occlusion, self-occlusion, and signal miss \cite{btcdet}. As a result, objects in point clouds can be broadly categorized into two groups: those with dense points and minimal occlusion (referred to as easy objects), and those with sparse points and significant occlusion (referred to as difficult objects).

\begin{figure}[!]
    \centering
    \resizebox{\linewidth}{!}{
    \includegraphics[width=1\columnwidth]{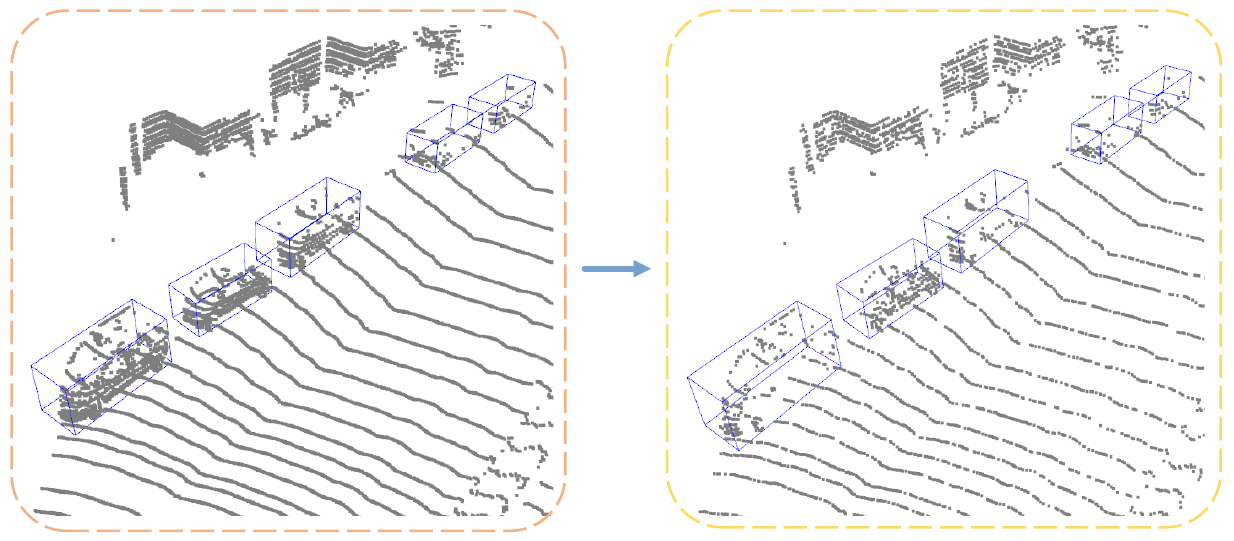}}
    \caption{A diagram of object resampling data augmentation that sparsifies points to different degrees in terms of distance and randomly drops points from the surfaces of easy objects. }
    \label{fig6}
\end{figure}

Since easy objects tend to be more complete and numerous in point clouds, 3D object detectors are typically more sensitive to these types of objects and exhibit higher detection accuracy for them. To alleviate this problem, we design a new object resampling data augmentation which undersamples easy objects and oversamples difficult objects. As shown in Figure~\ref{fig6}, it randomly transforms some easy objects into difficult objects, thus increasing the number and diversity of difficult objects. Extensive experiments show that our object resampling data augmentation can effectively improve the detection accuracy for difficult objects while not affecting the detection accuracy for easy objects.

Specifically, the object resampling data augmentation contains the following operations: (i) \textbf{\textit{sparsifing point cloud}} sets three ranges \{\textit{near, mid, far}\} in point clouds according to the distance from LiDAR. The sampling rates \{$p_1, p_2, p_3$\} of the three ranges decrease with distance from far to near. For simplicity and efficiency, we use random sampling instead of farthest point sampling \cite{pointnet++}. (ii) \textbf{\textit{random occlusion}} randomly selects some easy objects and removes points in one or two randomly chosen surfaces of these objects using the pyramid dropout method \cite{se-ssd}.

\begin{table*}[t]
    \caption{Performance comparisons on the KITTI \textit{test} set, evaluated by the average precision of 40 sampling recall points on the KITTI server. The best results of one-stage and two-stage detectors are highlighted in bold, respectively. ``-" indicates the related value is not given in the corresponding reference.}
    \centering
    \begin{tabular}{c|cccccccccccc}
    \hline
    & \multirow{2}{*}{Method} & \multirow{2}{*}{Modality} &\multicolumn{4}{c}{3D} && \multicolumn{4}{c}{BEV} & \multirow{2}{*}{\makecell[c]{Speed\\(ms)}}\\
    \cline{4-7} \cline{9-12}
                 &&&  Easy  & Mod.  &   Hard   &     mAP    &&  Easy  & Mod.  &   Hard   &     mAP &\\
    \hline
    \multirow{18}{*}{\begin{sideways} Two-stage \end{sideways}} &MV3D \cite{MV3D} & LiDAR+RGB & 74.97 & 63.63 & 54.00 & 64.20 && 86.62 & 78.93 & 69.80 & 78.45 & 360 \\
    &F-PointNet\cite{f-pointnet} & LiDAR+RGB & 82.19 & 69.79 & 60.59 & 70.86 && 91.17 & 84.67 & 74.77 & 83.54 & 170\\
    &AVOD \cite{avod} & LiDAR+RGB & 83.07 & 71.76 & 65.73 & 73.52 && 89.75 & 84.95 & 78.32 & 84.34 & 100\\
    &PointRCNN \cite{pointrcnn} & LiDAR & 86.96 & 75.64 & 70.70 & 77.77 && 92.13 & 87.39 & 82.72 & 87.41 & 100\\
    &F-ConvNet \cite{fconvnet} & LiDAR+RGB & 87.36 & 76.39 & 66.69 & 76.81 && 91.51 & 85.84 & 76.11 & 84.49 & 470\\
    &3D IoU Loss \cite{3diou} & LiDAR & 86.16 & 76.50 & 71.39 & 78.02 && 91.36 & 86.22 & 81.20 & 86.26 & 80 \\
    &Fast PointRCNN \cite{fastpr} & LiDAR & 85.29 & 77.40 & 70.24 & 77.64 && 90.87 & 87.84 & 80.52 & 86.41 & 65\\
    & UberATG-MMF \cite{mmf} & LiDAR+RGB & 88.40 & 77.43 & 70.22 & 78.68 && 93.67 & 88.21 & 81.99 & 87.96 & 80\\
    & Part-$A^2$ \cite{parta2} & LiDAR & 87.81 & 78.49 & 73.51 & 79.94 && 91.70 & 87.79 & 84.61 & 88.03 & 80\\
    & STD \cite{std} & LiDAR & 87.95 & 79.71 & 75.09 & 80.92 && 94.74 & 89.19 & 86.42 & 90.12 & 80\\
    & 3D-CVF \cite{3d-cvf} & LiDAR+RGB & 89.20 & 80.05 & 73.11 & 80.79 && 93.52 & 86.56 & 82.45 & 88.51 & 75\\
    & PV-RCNN \cite{pv-rcnn}& LiDAR & 90.25 & 81.43 & 76.82 & 82.83 && 94.98 & 90.65 & 86.14 & 90.59 & 80\\
    & BADet \cite{badet} & LiDAR & 89.28 & 81.61 & 76.58 & - && \textbf{95.23} & \textbf{91.32} & \textbf{86.48} & \textbf{91.01} & 140 \\ 
    & Voxel R-CNN \cite{voxel-rcnn} & LiDAR & \textbf{90.90} & 81.62 & 77.06 & 83.19 && 94.85 & 88.83 & 86.13 & 89.94 & \textbf{40}\\
    & ASCNet \cite{ascnet} & LIDAR & 88.48 & 81.67 & 76.93 & 82.36 && 92.85 & 89.36 & 86.45 & 89.55 &90 \\ 
    & SIENet \cite{sienet} & LiDAR & 88.22 & 81.71 & 77.22 & 82.38 && 92.38 & 88.65 & 86.03 & 89.02 &161 \\ 
    & BtcDet \cite{btcdet} & LiDAR &90.64 & \textbf{82.86} & \textbf{78.09} & \textbf{83.86} && 92.81 & 89.34 & 84.55 & 88.90 & 90\\
    \hline
    \multirow{16}{*}{\begin{sideways} Single-stage \end{sideways}} & VoxelNet \cite{voxelnet} & LiDAR & 77.82 & 64.17 & 57.51 & 66.5 && 87.95 & 78.39 & 71.29 & 79.21 & 220\\
    & ContFuse \cite{deepcf} & LiDAR+RGB & 83.68 & 68.78 & 61.67 & 71.38 && 94.07 & 85.35 & 75.88 & 85.10 & 60\\
    & SECOND \cite{second} & LiDAR & 83.34 & 72.55 & 65.82 & 73.90 && 89.39 & 83.77 & 78.59 & 83.92 & 50\\
    & PointPillars \cite{pointpillars} & LiDAR & 82.58 & 74.31 & 68.99 & 75.29 && 90.07 & 86.56 & 82.81 & 86.48 & 24\\
    & SMS-Net \cite{sms} & LiDAR & 87.01 & 76.21 & 70.45 & 77.89 && - & - & - & - & 24 \\ 
    & SVDNet \cite{svdnet} & LiDAR & 84.14 & 76.67 & 71.68 & 77.50 && - & - & - & - & - \\
    &Associate-3Ddet \cite{Associate3DdetPA} & LiDAR & 85.99 & 77.40 & 70.53 & 77.97 && 91.40 & 88.09 & 82.96 & 87.48 & 60\\
    &HotSpotNet \cite{hotspots} & LiDAR & 87.60 & 78.31 & 73.34 & 79.75 && 94.06 & 88.09 & 83.24 & 88.46 & 40\\
    & Point-GNN \cite{point-gnn} & LiDAR & 88.33 & 79.47 & 72.29 & 80.03 && 93.11 & 89.17 & 83.90 & 88.73 & 643\\
    & 3DSSD \cite{3dssd} & LiDAR & 88.36 & 79.57 & 74.55 & 80.83 && 92.66 & 89.02 & 85.86 & 89.18 & 38\\
    & SA-SSD \cite{sa-ssd} & LiDAR & 88.75 & 79.79 & 74.16 & 80.90 && 95.03 & 91.03 & 85.96 & 90.67 & 40\\
    & 3D-CenterNet \cite{3dcenternet} & LiDAR & 86.83 & 80.17 & 75.96 & 80.99 && 91.39 & 87.89 & 85.24 & 88.17 & -\\
    & CIA-SSD \cite{cia-ssd} & LiDAR & 89.59 & 80.28 & 72.87 & 80.91 && 93.74 & 89.84 & 82.39 & 88.60 & 31\\
    & IA-SSD \cite{ia-ssd} & LiDAR & 88.87 & 80.32 & 75.10 & 81.43 && 92.79 & 89.33 & 84.35 & 88.82 & \textbf{12}\\
    \cline{2-13}
    & NIV-SSD (ours) & LiDAR & \textbf{90.98} & \textbf{81.95} & \textbf{76.83} & \textbf{83.25} && \textbf{95.66} & \textbf{91.69} &  \textbf{86.72} &  \textbf{91.36} & 29\\
    \hline
\end{tabular}
\label{table1}
\end{table*}

\section{Experiments}
In this section, we evaluate models on widely-used 3D object detection benchmark datasets including KITTI \cite{kitti}, ONCE \cite{once} and Waymo Open \cite{waymo}. When evaluating models on the \textit{val} and \textit{test} sets of KITTI, we use the \textit{train} set and the union of the \textit{train} and \textit{val} sets for training, respectively. On the ONCE dataset, we use the official splits to train and evaluate models. As for the Waymo Open dataset, following \cite{pv-rcnn}, 20\% samples from the \textit{train} set are used for training.  

\subsection{Implementation Setup}

\noindent\textbf{Data preprocessing.} 
The detection range and voxel size on KITTI, ONCE, and Waymo Open datasets are kept the same as \cite{second}, \cite{once} and \cite{pv-rcnn}, respectively. For the object resampling data augmentation, we empirically set \textit{near} = [0, 20), \textit{mid} = [20, 35), and \textit{far} = [35, $+\infty$), and $p_1$, $p_2$, and $p_3$ are randomly sampled from ranges [0.4, 0.6], [0.6, 0.8], and [0.8, 1.0], respectively. The easy objects drop points on \{0, 1, 2\} surfaces with probabilities of \{0.25, 0.5, 0.25\}, respectively. Besides the object resampling data augmentation, we adopt the following data augmentations: (i) ground-truth augmentation \cite{second}; (ii) global augmentations including random flipping, rotation, and scaling on a whole point cloud. The global rotation augmentation used in our NIV-SSD is around the X, Y, and Z axes, and the rotation angles are randomly sampled from ranges [-0.035, 0.035], [-0.025, 0.025], [-0.785, 0.785], respectively. Rotation around the X and Y axes is to simulate the situation of ground tilt; (iii) local augmentations including random rotation and translation on local ground truths; (iv) similar category filtering \cite{cia-ssd} treats objects of similar categories as the objects of target categories, such as van for car, to alleviate model confusion in training; (v) shape-aware augmentation \cite{se-ssd}.

\setlength{\parskip}{1ex}

\noindent\textbf{Training Details.}
All models are trained from scratch in an end-to-end manner with the AdamW optimizer \cite{adamw} and one-cycle policy \cite{one_cycle} with a learning rate of 0.001. The \textit{score\_thres}, and \textit{iou\_thres} in the neighbor IoU-voting strategy are empirically set to 0.1 and 0.2, respectively. On the KITTI, ONCE and Waymo Open datasets, models are trained for 60 epochs with a batch size of 8, 80 epochs with a batch size of 8, and 30 epochs with a batch size of 16, respectively.  

\setlength{\parskip}{1ex}

\subsection{Comparisons on the KITTI Dateset}

\begin{table*}[t]
\caption{Performance comparisons on the KITTI \textit{val} set, evaluated by AP under 40 sampling recall points (R40) and 11 sampling recall points (R11). ``*" represents that the method is re-implemented using the same data augmentations with NIV-SSD. ``-" indicates the related value is not given in the corresponding reference.}
\centering
\begin{tabular}{cccccccccccc}
    \hline
    \multirow{2}{*}{Method} & \multirow{2}{*}{Modality} & \multirow{2}{*}{Stage} &\multicolumn{4}{c}{Car 3D $AP_{R40}$} &&  \multicolumn{4}{c}{Car 3D $AP_{R11}$}\\
    \cline{4-7} \cline{9-12}
    &&&  Easy  & Mod.  &   Hard  & mAP && Easy & Mod. & Hard & mAP \\
    \hline
    VoxelNet \cite{voxelnet} & LiDAR & One & -   &   -   &   -   &   -   && 81.97 & 65.46 & 62.85 & 70.09 \\
    ContFuse \cite{deepcf} & LiDAR+RGB & One & -   &   -   &   -   &   -   && 86.32 & 73.25 & 67.81 & 75.79 \\
    
    3D-CenterNet \cite{3dcenternet} & LiDAR & One & 92.14 & 82.93 & 80.76 & 84.61 && - & - & - & -\\
    SVDNet \cite{svdnet} & LiDAR & One & - & - & - & - && 88.21 & 77.72 & 75.55 & 80.49 \\
    SMS-Net \cite{sms}   & LiDAR & One & - & - & - & - && 89.34 & 79.04 & 77.76 & 82.05 \\
    ASCNet \cite{ascnet} & LiDAR & One & - & - & - & - && 89.12 & 79.25 & 78.58 & 82.32 \\
    CIA-SSD \cite{cia-ssd} & LiDAR & One &  -   &   -   &   -   &   -   && 90.04 & 79.81 & 78.80 & 82.88 \\
    SA-SSD \cite{sa-ssd}  & LiDAR & One & 92.23 & 84.30 & 81.36 & 85.96 && \textbf{90.15} & 79.91 & 78.78 & 82.95\\
    
    PV-RCNN  \cite{pv-rcnn} & LiDAR & Two & 92.57 & 84.83 & 82.69 & 86.70 &&   -   & 83.90 &   -   & -   \\
    Voxel R-CNN \cite{voxel-rcnn} & LiDAR & Two & 92.38 & 85.29 & 82.86 & 86.84 && 89.41 & 84.52 & 78.93 & 84.29 \\
    SIENet \cite{sienet}& LiDAR & Two & 92.49 & 85.43 & 83.05 & 86.99 && - & 84.40 & - & - \\
    BAdet \cite{badet} & LiDAR & Two & - & - & - & - && 90.06 & 85.77 & 79.00 & 84.93 \\
    BtcDet \cite{btcdet} & LiDAR & Two & 93.15 & 86.28 &\textbf{83.86}& 87.76 &&   -   & \textbf{86.57} & - & -\\
    \hline
    PointPillars* \cite{pointpillars} & LiDAR & One & 91.49 & 80.06 & 78.99 & 83.51 && 88.66 & 78.22& 77.10& 81.32\\
    PointPillars* w/ \textit{NIV} & LiDAR & One & 92.37 & 80.60 & 79.53 & 84.17 && 89.22 & 78.61& 77.51& 81.78\\
    \rowcolor{lgray} Improvement $\uparrow$ &  N/A & N/A & +0.88 & +0.54 & +0.54 & +0.66&& +0.56& +0.39& +0.41&+0.46\\
    \hline
    SECOND* \cite{second} & LiDAR & One & 93.09 & 85.17 & 82.12 & 86.79 && 89.88 & 84.90 & 78.20 & 84.32 \\
    SECOND* w/ \textit{NIV} & LiDAR & One & 93.23 & 85.71 & 82.77 & 87.23 && 89.83 & 85.79& 78.56 & 84.73\\
    \rowcolor{lgray} Improvement $\uparrow$ & N/A & N/A & +0.14 & +0.54 & +0.65 & +0.44 && -0.05 & +0.89 & +0.36& +0.41\\
    \hline
    NIV-SSD (ours) & LiDAR & One &  \textbf{93.58} & \textbf{86.41} & 83.43 & \textbf{87.81} && \textbf{90.15} & 86.39 & 79.05 & \textbf{85.20} \\
    \hline
\end{tabular}
\label{table2}
\end{table*}

\noindent\textbf{3D Detection.}
We submit the prediction results of our NIV-SSD on the KITTI \textit{test} set to the online KITTI server\footnote{https://www.cvlibs.net/datasets/kitti}. As depicted in Table~\ref{table1}, our NIV-SSD achieves the best performance in terms of 3D detections on all metrics among the SOTA single-stage detectors. Note that the ``moderate AP" is the official ranking metric of the KITTI dataset. Our NIV-SSD outperforms the state-of-the-art single-stage methods greatly on the ``moderate AP" metric. Generally, two-stage detectors perform better than single-stage detectors due to their extra refinement. Despite that, our NIV-SSD still outperforms most of the two-stage detectors and achieves results close to the SOTA two-stage method BtcDet. In addition, NIV-SSD can run at the speed of 29 ms per example on a single 3090 GPU, which is much faster than most two-stage detectors. Table~\ref{table2} shows the results of our NIV-SSD and several state-of-the-art methods on the KITTI \textit{val} set. As we can see, NIV-SSD surpasses most of the state-of-the-art methods and even performs better than BtcDet on easy and moderate levels. Additionally, the performances of PointPillars \cite{pointpillars} and SECOND \cite{second} are greatly improved by our \textit{NIV} strategy, demonstrating the effectiveness of this strategy. 

Note that the \textit{NIV} is a post-processing method, so it can be directly plugged into a trained single-stage detector. Our NIV-SSD slightly performs better on the KITTI \textit{val} set. As mentioned in prior works \cite{parta2, cia-ssd}, such difference may be caused by the inconsistency distribution between the KITTI \textit{test} and \textit{val} sets. 



\noindent\textbf{BEV Detection.}
Table~\ref{table1} presents the results of our BEV detection experiments, revealing that our NIV-SSD model surpasses all single-stage and two-stage detectors on different detection levels. Interestingly, we observed that the advantage of two-stage detectors over single-stage detectors in BEV detection is not as pronounced as it is in 3D detection. We posit that this phenomenon arises from the fact that two-stage detectors can utilize fine-grained height information from 3D feature maps to refine 3D bounding boxes, whereas single-stage detectors are typically limited to using compressed 2D feature maps for this task.


\begin{table*}[t]
    \caption{Performance comparisons on the ONCE \textit{validation} set. The best results of detectors are highlighted in bold. ``*" represents that the method is re-implemented using the official code \cite{once}. }
    \centering
    \resizebox{\linewidth}{!}{
    \begin{tabular}{cccccccccccccccc}
    \hline
    \multirow{2}{*}{Method} & \multicolumn{4}{c}{Vehicle} && \multicolumn{4}{c}{Pedestrian} && \multicolumn{4}{c}{Cyclist} & \multirow{2}{*}{mAP}\\
    \cline{2-5} \cline{7-10} \cline{12-15}
    & overall & 0-30m & 30-50m & $>$50m && overall & 0-30m & 30-50m & $>$50m && overall & 0-30m & 30-50m & $>$50m &\\
    \hline
    PointRCNN \cite{pointrcnn} & 52.09 & 74.45 & 40.89 & 16.81 && 4.28 & 6.17 & 2.40 & 0.91 && 29.84 & 46.03 & 20.94 & 5.46 & 28.74 \\
    Centerpoint \cite{centerpoint} & 66.79 & 80.10 & 59.55 & 43.39 && \textbf{49.90} & \textbf{56.24} & \textbf{42.61} & \textbf{26.27} && 63.45 & 74.28 & 57.94 & 41.48 & 60.05 \\
    IA-SSD \cite{ia-ssd} & 70.30 & 83.01 & 62.84 & 47.01 && 39.82 & 47.45 & 32.75 & 18.99 && 62.17 & 73.78 & 56.31 & 39.53 & 57.43 \\
    PV-RCNN \cite{pv-rcnn} & 77.77 &\textbf{89.39} & 72.55 & 58.64 && 23.50 & 25.61 & 22.84 & 17.27 && 59.37 & 71.66 & 52.58 & 36.17 & 53.55 \\
    \hline
    Pointpillars* \cite{pointpillars} & 70.56 & 82.56 & 64.18 & 50.98 && 20.43 & 22.98 & 18.17 & 11.06 && 53.10 & 63.72 & 47.48 & 31.57 & 48.03 \\
    Pointpillars* w/ \textit{NIV} & 71.95 & 83.53 & 64.84 & 51.53 && 20.50 & 23.06 & 18.17 & 11.53 && 53.38 & 64.10 & 47.50 & 31.88 & 48.61 \\
    \rowcolor{lgray} Improvement $\uparrow$ & +1.39 & +0.97 & +0.66 & +0.55 && +0.07 & +0.08 & 0.0 & +0.47 && +0.28 & +0.38 & +0.02 & +0.31 & +0.58 \\
    \hline
    SECOND* \cite{second}& 75.08 & 85.17 & 70.48 & 56.79 && 31.38 & 35.05 & 27.87 & 20.26 && 61.74 & 72.28 & 56.61 & 39.87 & 56.07 \\
    SECOND* w/ \textit{NIV} & 75.95 & 86.26 & 71.27 & 57.50 && 31.45 & 35.05 & 28.05 & 20.69 && 61.83 & 72.47 & 56.75 & 39.77 & 56.41 \\
    \rowcolor{lgray} Improvement $\uparrow$ & +0.87 & +1.09 & +0.79 & +0.71 && +0.07 & +0.00 & +0.18 & +0.43 && +0.09 & +0.19 & +0.14 & -0.10 & +0.34 \\
    \hline
    NIV-SSD (ours) & \textbf{78.31} & 87.32 & \textbf{72.84} & \textbf{59.51} && \textbf{37.22} & 41.40 & 33.55 & 24.50 && \textbf{65.65} & \textbf{76.18} & \textbf{60.31} & \textbf{43.48} & \textbf{60.39} \\
    \hline
\end{tabular}
}
\label{table3}
\end{table*}

\begin{table*}[t]
    \caption{Performance comparisons on the Waymo \textit{validation} set. The best results of detectors are highlighted in bold. ``*" represents that the method is re-implemented using the official codebase OpenPCDet \cite{openpcdet}.}
    \centering
    \resizebox{\linewidth}{!}{
    \begin{tabular}{cccccccccccccc}
    \hline
    \multirow{3}{*}{Method} & \multicolumn{6}{c}{LEVEL 1} & & \multicolumn{6}{c}{LEVEL 2} \\
    \cline{2-7} \cline{9-14}
    &\multicolumn{2}{c}{Vehicle} & \multicolumn{2}{c}{Pedestrian} & \multicolumn{2}{c}{Cyclist} & &\multicolumn{2}{c}{Vehicle} & \multicolumn{2}{c}{Pedestrian} & \multicolumn{2}{c}{Cyclist}\\
    \cline{2-7} \cline{9-14}
    & mAP & mAPH & mAP & mAPH & mAP & mAPH & &mAP & mAPH & mAP & mAPH & mAP & mAPH\\
    \hline
    Part-$A^2$ \cite{parta2} & 71.82 & 71.29 & 63.15 & 54.96 & 65.23 & 63.92 & &64.33 & 63.82 & 54.24 & 47.11 & 62.61 & 61.35 \\
    PV-RCNN \cite{pv-rcnn} & \textbf{74.06} & \textbf{73.38} & 62.66 & 52.68 & 63.32 & 61.71 && 64.99 & 64.38 & 53.80 & 45.14 & 60.72 & 59.18  \\
    \hline
    Pointpillars* \cite{pointpillars} & 67.07 & 66.37 & 60.91 & 39.46 & 52.70 & 48.35 & &59.00 & 58.37 & 53.37 & 34.51 & 50.77 & 46.58 \\
    Pointpillars* w/ \textit{NIV} & 67.55 & 66.85 & 64.03 & 41.71 & 53.03 & 48.92 & &59.43 & 58.79 & 56.08 & 36.45 & 51.08 & 47.13 \\
    \rowcolor{lgray} Improvement $\uparrow$ & +0.48 & +0.48 & +3.12 & +2.25 & +0.33 & +0.57 && +0.43 & +0.42 & +2.71 & +1.94 & +0.31 & +0.55 \\
    \hline
    SECOND* \cite{second} & 70.67 & 70.09 & 67.72 & 58.24 & 61.10 & 59.58 & & 62.52 & 61.99 & 59.44 & 50.95 & 58.80 & 57.33 \\
    SECOND* w/ \textit{NIV} & 71.01 & 70.43 & 68.43 & 58.68 & 61.61 & 60.10 & &62.84 & 62.31 & 60.01 & 51.30 & 59.28 & 57.83 \\
    \rowcolor{lgray} Improvement $\uparrow$ & +0.34 & +0.34 & +0.71 & +0.44 & +0.51 & +0.52 & &+0.32 & +0.32 & +0.57 & +0.35 & +0.48 & +0.50 \\  
    \hline
    NIV-SSD (ours)& 73.66 & 73.11 & \textbf{72.09} & \textbf{63.59} & \textbf{66.09} & \textbf{64.83} & &\textbf{65.28} & \textbf{64.77} & \textbf{62.89} & \textbf{55.30} & \textbf{63.66} & \textbf{62.44} \\
    \hline
\end{tabular}
}
\label{table4}
\end{table*}

\subsection{Comparisons on the ONCE Dataset}

To comprehensively demonstrate the effectiveness and generality of our \textit{NIV} strategy and NIV-SSD detector, we conducted experiments on the ONCE \cite{once} dataset. Table~\ref{table3} presents the results, revealing that the \textit{NIV} strategy enhances the performance of PointPillars \cite{pointpillars} and SECOND \cite{second} across different categories, with a particular improvement in the ``Vehicle" category. Additionally, we observed that our NIV-SSD model achieves the best results across most metrics, demonstrating the effectiveness of the NIV-SSD.

\subsection{Comparisons on the Waymo Dataset}

We have conducted further experiments on the Waymo \cite{waymo} dataset to validate the effectiveness of our \textit{NIV} strategy and NIV-SSD detector. The results in Table~\ref{table4} demonstrate that our \textit{NIV} strategy significantly improves the performance of PointPillars \cite{pointpillars} and SECOND \cite{second} on all evaluation metrics. Furthermore, our NIV-SSD detector is shown to be a competitive baseline for single-stage detectors on the Waymo dataset.

\subsection{Ablation Study}

To further study the influence of each component of NIV-SSD, we perform a comprehensive ablation analysis on the KITTI dataset. All models are trained on \textit{train} set and evaluated on \textit{val} set. Table~\ref{table5},~\ref{table6},~\ref{table7} show the effect of the proposed modules including the object resampling data augmentation (\textit{OR-DA}), lite 3DSparseResNet (\textit{L-RES}), ConvNeXt block (\textit{CN}) and neighbor IoU-voting strategy (\textit{NIV}).

\noindent\textbf{Effect of \textit{OR-DA}.} We utilized SECOND as the baseline model without any data augmentations. The accuracy of the model increased with each data augmentation scheme employed, as presented in Table~\ref{table5}. Notably, the \textit{OR-DA} technique proved to be effective in enhancing the performance of the baseline model across all difficulty levels, especially for moderate and hard levels, as shown in the 5th and 6th rows of Table~\ref{table5}. These findings suggest that the \textit{OR-DA} technique, which undersamples easy objects and oversamples difficult objects by randomly transforming easy objects into difficult objects, can effectively address the imbalance in detection accuracy between easy and difficult point cloud objects. 

\begin{table}[t]
    \caption{Effect of different data augmentation methods. The 3D average precisions of 40 sampling recall points on KITTI \textit{val} set for car detection are reported. \textit{GLOBAL}, \textit{LOCAL}, \textit{GT}, \textit{SIM}, \textit{SA}, and \textit{OR} denote global augmentations, local augmentations, ground-truth augmentation, similar category filtering, shape-aware augmentation, and object resampling augmentation, respectively.}
    \centering
    \resizebox{\linewidth}{!}{
    \begin{tabular}{ccccccccc}
        \hline
        \textit{GLOBAL} & \textit{LOCAL} & \textit{GT}  & \textit{SIM} & \textit{SA} &\textit{OR} & Easy & Mod. &  Hard\\
        \hline
        &&&&&&71.85 & 64.06 & 60.09\\
        $\checkmark$&&&&&&88.84 & 79.40 & 76.70\\
        $\checkmark$&$\checkmark$&&&&& 91.15	& 80.40 & 77.45\\
        $\checkmark$&$\checkmark$&$\checkmark$&&&& 91.80  & 82.82 & 79.48\\
        $\checkmark$&$\checkmark$&$\checkmark$&$\checkmark$&&& 92.45	& 83.42 & 80.01\\
        $\checkmark$&$\checkmark$&$\checkmark$&$\checkmark$&$\checkmark$&& 92.98	& 84.13 & 80.76\\
        $\checkmark$&$\checkmark$&$\checkmark$&$\checkmark$&$\checkmark$& $\checkmark$ & 93.09 & 85.17 & 82.12\\
        \hline
    \end{tabular}}
\label{table5}
\end{table}

\begin{table}[t]
    \caption{Effect of our proposed modules. The 3D average precisions of 40 sampling recall points on KITTI \textit{val} set for car detection are reported. }
    \centering
    \begin{tabular}{cccc}
        \hline
        Methods & Easy & Mod. &  Hard\\
        \hline
        baseline & 93.09 & 85.17 & 82.12\\
        baseline w/ \textit{L-RES} &93.06& 85.43& 82.26\\
        baseline w/ \textit{CN} &93.02& 85.30& 82.32\\
        baseline w/ \textit{L-RES}, \textit{CN} & 93.09& 85.48& 82.39\\
        \hline
    \end{tabular}
\label{table6}
\end{table}

\begin{figure*}[t]
    \centering
    \includegraphics[width=0.95\textwidth]{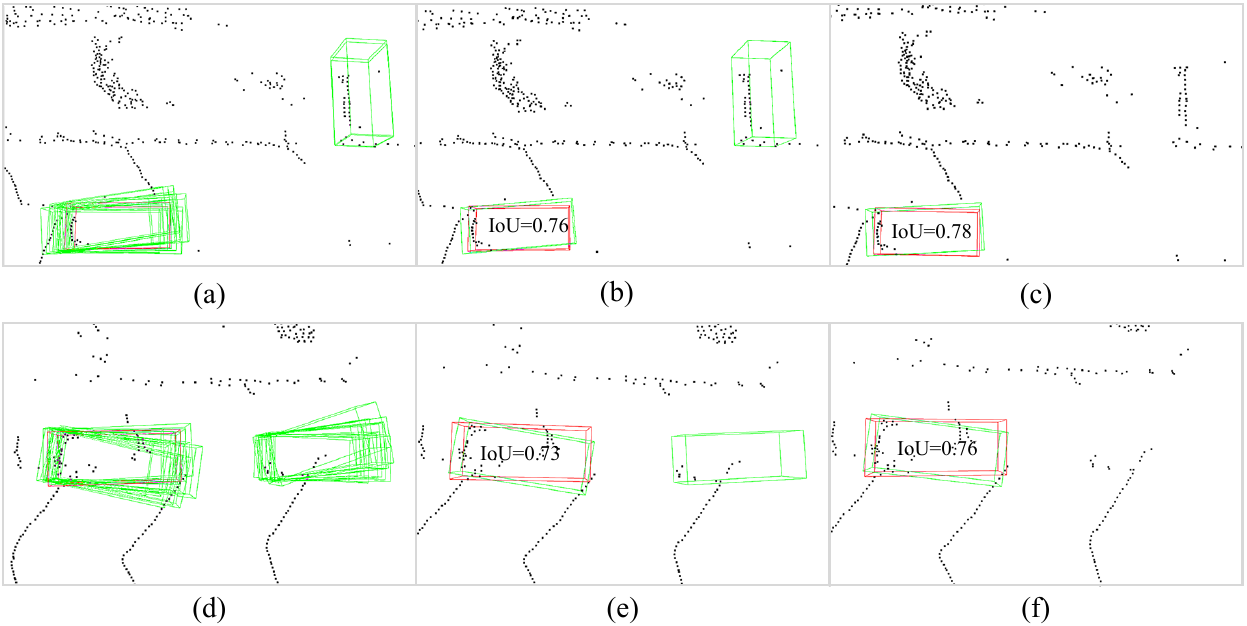}
    \caption{Visualization of prediction results without NMS, with NMS, and with our NIV and NMS, respectively. (a) and (d) show results without NMS. (b) and (e) show results with NMS. (c) and (f) show results with our NIV and NMS. The predicted and ground-truth bounding boxes are shown in green and red, respectively.}
    \label{fig5}
\end{figure*}

\noindent\textbf{Effect of \textit{L-RES} and \textit{CN}.} The baseline model employed here is SECOND with aforementioned data augmentation techniques. The results presented in the first and second rows of Table~\ref{table6} reveal that \textit{L-RES} surpasses the performance of the 3D backbone utilized in SECOND. Furthermore, the replacement of traditional convolution layers with \textit{CN} resulted in an increase in APs across all levels of difficulty, as can be seen in the first and third rows of Table~\ref{table6}. The combination of \textit{L-RES} and \textit{CN} further improved the detection accuracy, as demonstrated in the fourth row of Table~\ref{table6}. These results strongly suggest that both \textit{L-RES} and \textit{CN} significantly enhance the feature extraction capabilities of the model.

\begin{table}[t]
    \caption{Effect of our proposed \textit{NIV} strategy. The 3D average precisions of 40 sampling recall points on KITTI \textit{val} set for car detection are reported. }
    \centering
    \begin{tabular}{cccc}
        \hline
        Methods & Easy & Mod. &  Hard\\
        \hline
        baseline &93.09& 85.48& 82.39\\
        baseline w/ \textit{NIV} & 93.29& 86.00& 82.88\\
        \rowcolor{lgray} Improvement $\uparrow$ & +0.20 & +0.52 & +0.49\\
        \hline
        baseline w/ QFL & 92.56& 85.71 & 83.22\\
        baseline w/ QFL, \textit{NIV} & 92.70& 85.99 & 83.48\\
        \rowcolor{lgray} Improvement $\uparrow$ & +0.14& +0.28 & +0.26\\
        \hline
        baseline w/ IoU-aware & 93.54 & 86.04 & 83.13\\
        baseline w/ IoU-aware, \textit{NIV} & 93.58 & 86.41 & 83.43\\
        \rowcolor{lgray} Improvement $\uparrow$ & +0.04 & +0.37 & +0.30\\
        \hline
    \end{tabular}
\label{table7}
\end{table}

\begin{figure}[t]
    \centering
    \includegraphics[width=1\columnwidth]{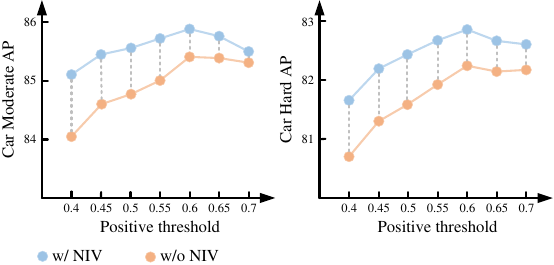}
    \caption{Effect of \textit{NIV} on different positive and negative thresholds. Negative thresholds are 0.15 lower than corresponding positive thresholds. The 3D average precisions of 40 sampling recall points on KITTI \textit{val} set for car detection are reported. }
    \label{fig8}
\end{figure}

\noindent\textbf{Effect of \textit{NIV}.} SECOND with aforementioned data augmentations, \textit{L-RES}, and \textit{CN} is acted as the baseline model. As shown in Figure~\ref{fig8}, the adoption of the \textit{NIV} strategy yields a discernible enhancement in detection accuracy across varying positive and negative thresholds, with particularly pronounced gains observed when such thresholds are set to lower values. We contend that lower positive thresholds lead to more neighbors for an object, thus the \textit{NIV} strategy can leverage more precise statistical data from neighbors. By setting the positive and negative thresholds at 0.6 and 0.45, respectively, the detector attains a high level of performance. Hence, we employ these values as the default positive and negative thresholds.

\begin{figure*}[t]
    \centering
    \includegraphics[width=0.95\textwidth]{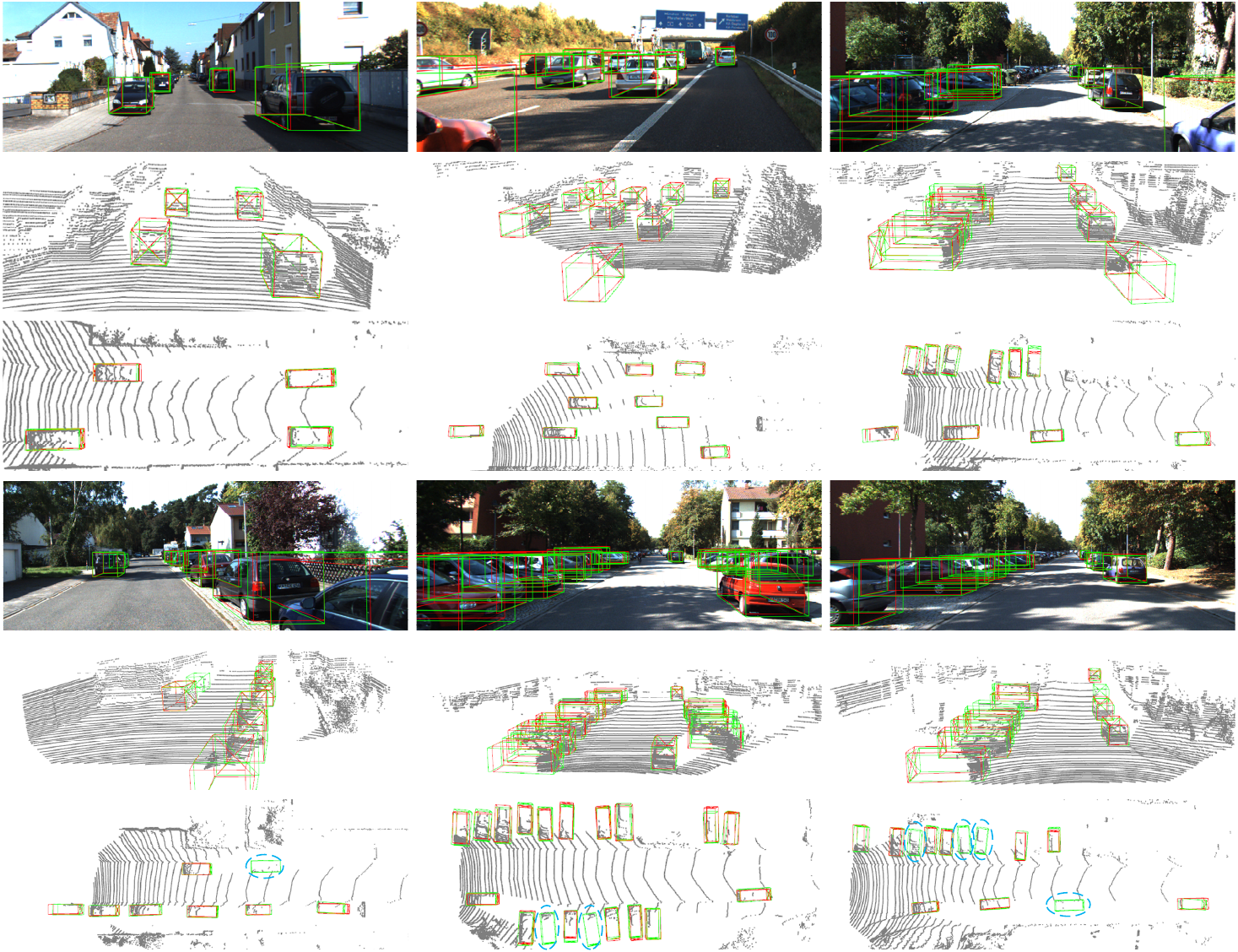}
    \caption{Visualization of 3D and BEV detection results of NIV-SSD on KITTI validation set. The ground-truth and predicted bounding boxes are projected back to images and rendered in red and green, respectively. Blue circles indicate missing ground truth bounding boxes. }
    \label{fig7}
\end{figure*}

Table~\ref{table7} demonstrates that our \textit{NIV} strategy significantly improves the baseline model, particularly for objects of moderate and hard difficulty levels (1st and 2nd rows). Additionally, the IoU-aware and quality focal loss also enhance the detection accuracy (4th and 7th rows). It is worth noting that the \textit{NIV} strategy can be combined with other confidence rectification techniques, such as the IoU-aware and quality focal loss. By combining these methods (5th and 8th rows of Table~\ref{table7}), the \textit{NIV} strategy further enhances the detection accuracy, especially for objects of moderate and hard levels. These findings indicate that the \textit{NIV} strategy effectively rectifies the classification confidence for objects of moderate and hard levels, while having minimal impact on objects of easy level. This may be due to that easy objects are relatively stable (i.e., predicted bounding boxes for an object are close to overlapping), thus good-quality predictions are not easy to filter out by NMS. In contrast, for moderate and hard objects, the predicted bounding boxes are more unstable (i.e., predicted bounding boxes for an object may vary greatly), making it easier to eliminate good-quality predictions and retain poor-quality ones. 

\subsection{Qualitative Analysis about NIV}

To comprehensively clarify how the proposed \textit{NIV} works, we show some prediction results of NIV-SSD in Figure~\ref{fig5}. There are false positive predictions in Figure~\ref{fig5}(a), they cannot be filtered out using only NMS as shown in Figure~\ref{fig5}(b). Utilizing the number of neighbors, as shown in Figure~\ref{fig5}(c), our \textit{NIV} can eliminate these redundant predictions. As presented in Figure~\ref{fig5}(d), the false positive predictions are very unstable. They also cannot be filtered out using NMS as shown in Figure~\ref{fig5}(e). The \textit{NIV} strategy can remove these redundant predictions using the mean IoU statistical data as shown in Figure~\ref{fig5}(f). And our \textit{NIV} can retain relatively good-quality bounding boxes from true positive predictions. As shown in Figure~\ref{fig5}, the IoUs between the final predicted and ground-truth bounding boxes are increased after applying the \textit{NIV} strategy. 

Figure~\ref{fig7} depicts the results predicted by our NIV-SSD from various views. The 2nd and 5th rows display the bounding boxes from a 3D view, while the 3rd and 6th rows illustrate the bounding boxes from a bird's eye view. The 3D bounding boxes are then projected back onto the images, as shown in the 1st and 4th rows. The results presented in the 1st, 2nd, and 3rd rows of Figure~\ref{fig7} evince the NIV-SSD's high detection accuracy. Moreover, as evinced in the 6th row of Figure~\ref{fig7}, ground-truth bounding boxes are missed for some objects. These overlooked objects contain minimal points in the point cloud, nonetheless, our NIV-SSD can still recognize and localize them.

\subsection{Quantitative Analysis about NIV}

In this section, we quantify the role of our \textit{NIV}, IoU-aware \cite{cia-ssd}, and the combination of the two methods. The experimental results are obtained from the ONCE \textit{validation} dataset. As presented in Table~\ref{table8}, both the \textit{NIV} score and predicted IoU improve the average precision (AP) and PCC values. After integrating the \textit{NIV} score and predicted IoU to the classification confidence score, the PCC and AP values can be further improved. It demonstrates that our \textit{NIV} can be combined with the IoU-aware method to further rectify the confidence score. We also observe that the rectified confidence scores are still far from the real IoUs between predicted boxes and ground-truth boxes, which limits the performance of detectors. We leave the improvement in future work.

\subsection{Model Size and Runtime Analysis}

\begin{table}
    \caption{Results of different confidence scores on the ONCE dataset. PCC denotes the Pearson correlation coefficient between real IoUs and confidence scores. ``CS",  ``NIV", ``pIoU" and ``rIoU" denote the classification confidence score, \textit{NIV} score, predicted IoU, and real IoU, respectively. }
    \centering
    \resizebox{\linewidth}{!}{
        \begin{tabular}{cccccc}
            \hline
            Method & CS & CS,pIoU & CS,NIV & CS,NIV,pIoU & rIoU \\
            \hline
            $\mathrm{AP}_{vehicle}$ & 77.09 & 77.92 & 77.98 & 78.31 & 86.39 \\
            PCC & 0.602 & 0.615 & 0.618 & 0.624 & 1.0\\
            \hline
        \end{tabular}
    }
\label{table8}
\end{table}

\begin{table}[t]
    \caption{Comparison of our NIV-SSD with baseline models on the number of parameters, runtime, and average precision on the KITTI \textit{val} set.}
    \centering
    \resizebox{\linewidth}{!}{
    \begin{tabular}{cccccc}
        \hline
        \multirow{2}{*}{Model} & \multirow{2}{*}{Params} & \multirow{2}{*}{Time (ms)} & \multicolumn{3}{c}{Car 3D $AP_{R40}$}\\
        \cline{4-6}
        &&& Easy  & Mod.  &   Hard \\
        \hline
        SECOND & 5.7M & 25.0 & 92.85 & 85.47 & 82.68 \\
        CIA-SSD & 3.6M & 23.4 & 93.43 & 85.51 & 82.75\\
        NIV-SSD & 3.4M & 28.5 & 93.58 & 86.41 & 83.43\\
        NIV-SSD w/o \textit{NIV} & 3.4M & 28.3 & 93.54 & 86.04 & 83.13\\
        \hline
    \end{tabular}}
\label{table9}
\end{table}

In this section, we compare the parameter number and runtime between our NIV-SSD and several baseline models including SECOND \cite{second} and CIA-SSD \cite{cia-ssd}. We re-implement SECOND and CIA-SSD and train them using the same data augmentation schemes as our NIV-SSD. All experiments are conducted on a single RTX3090 GPU. As Table~\ref{table9} demonstrates, NIV-SSD provides a well-balanced trade-off between speed and accuracy, enhancing the accuracy in multiple metrics with only a minor increase in latency compared to the baseline models. As the 1st and 4th rows of Table~\ref{table9} indicate, the \textit{NIV} strategy results in only a 0.2 ms latency while notably improving detection accuracy for moderate and hard levels. Moreover, the parameter number of NIV-SSD is minimal, which is also essential for memory-constrained devices.

\section{Conclusion}
In this paper, a single-stage object detector named neighbor IoU-voting single-stage object detector (NIV-SSD) is proposed. To solve the misalignment problem, we propose the \textit{NIV} strategy which utilizes two types of statistical data from regression output to rectify classification confidence, thereby establishing a connection between independent classification and regression branches. Furthermore, we introduce the object resampling data augmentation to balance the detection accuracy for easy and difficult objects. Combining the \textit{NIV} strategy and object resampling augmentation, we design a single-stage detector NIV-SSD with both speed and accuracy. Extensive experiments conducted on several datasets demonstrate the generality and effectiveness of the \textit{NIV} strategy and the superior performance of the NIV-SSD detector.

\bibliographystyle{elsarticle-num} 
\bibliography{ref}

\subsection*{  } 
\setlength\intextsep{0pt}
\begin{wrapfigure}{l}{25mm} 
\includegraphics[width=1in,height=1.25in,clip,keepaspectratio]{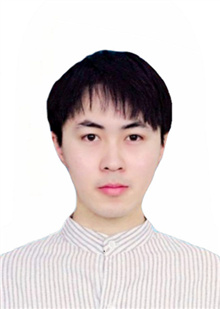}
\end{wrapfigure}\par
\noindent \textbf{Shuai Liu} received the B.Sc. degree in computer science and technology from Xidian University in 2020, the M.Sc. degree in computer technology from Xidian University in 2023. He is currently pursuing the Ph.D. degree with the School of Computer Science and Engineering at Sun Yat-sen University. His research interests focus on machine learning and computer vision.\par

\subsection*{  } 
\setlength\intextsep{0pt}
\begin{wrapfigure}{l}{25mm} 
{\includegraphics[width=1in,height=1.25in,clip,keepaspectratio]{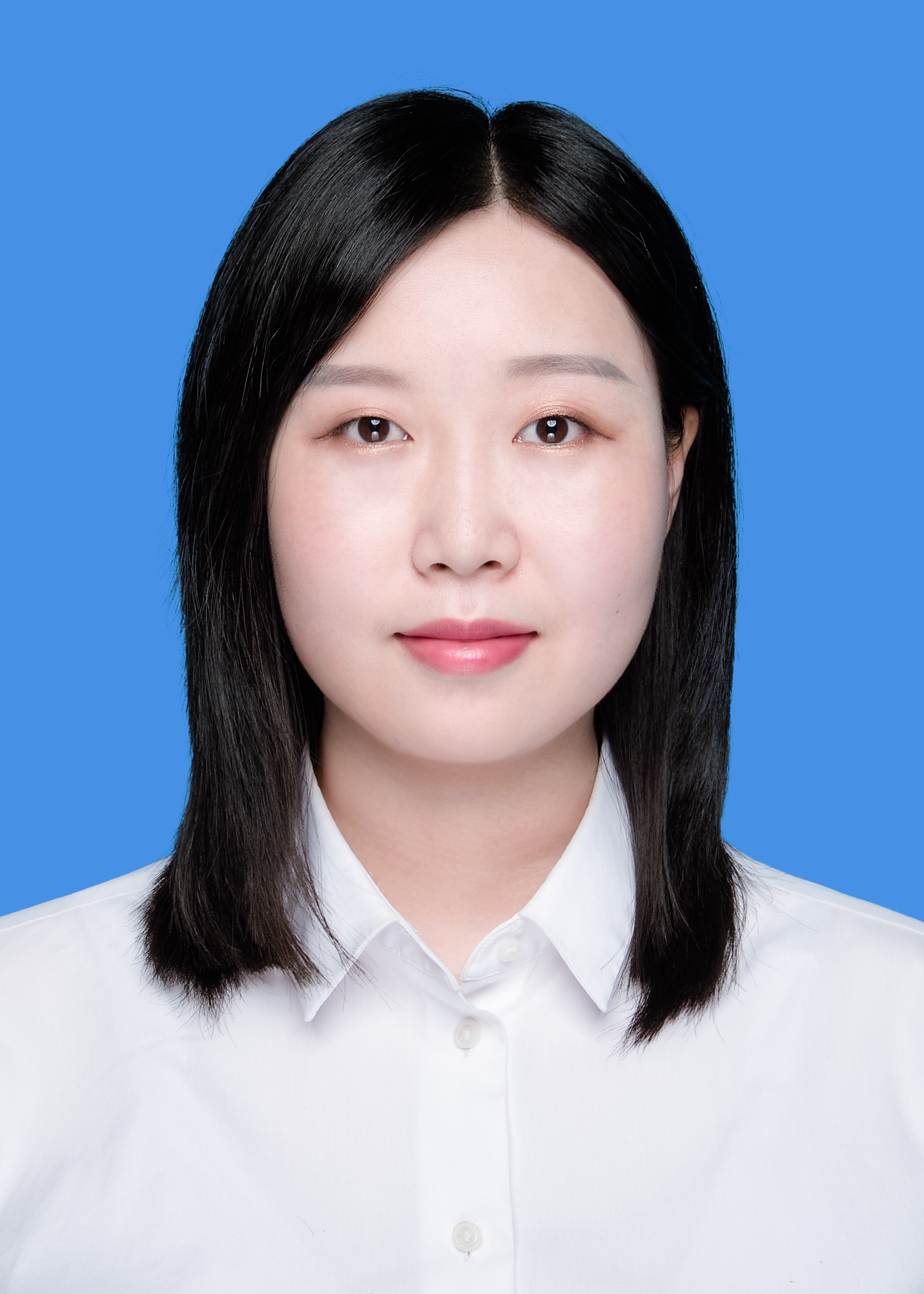}}
\end{wrapfigure}\par
\noindent \textbf{Di Wang} received the Ph.D. degree in intelligent information processing from Xidian University, Xi'an, China, in 2016. She is currently an Associate Professor in the School of Computer Science and Technology at Xidian University. Her research interests include machine learning and multimedia information retrieval. In these areas, she has published several scientific articles in refereed journals including the IEEE TPAMI, TIP, TCYB and TCSVT, and conferences including the SIGIR and IJCAI.\par

\subsection*{  } 
\setlength\intextsep{0pt}
\begin{wrapfigure}{l}{25mm} 
{\includegraphics[width=1in,height=1.25in,clip,keepaspectratio]{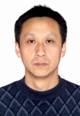}}
\end{wrapfigure}\par
\noindent \textbf{Quan Wang} received the B.Sc., M.Sc., and Ph.D. degrees in computer science and technology from Xidian University, Xi'an, China. He is currently a professor in the School of Computer Science and Technology at Xidian University. His current research interests include input and output technologies and systems, image processing and image understanding.\par

\subsection*{  } 
\setlength\intextsep{0pt}
\begin{wrapfigure}{l}{25mm} 
{\includegraphics[width=1in,height=1.25in,clip,keepaspectratio]{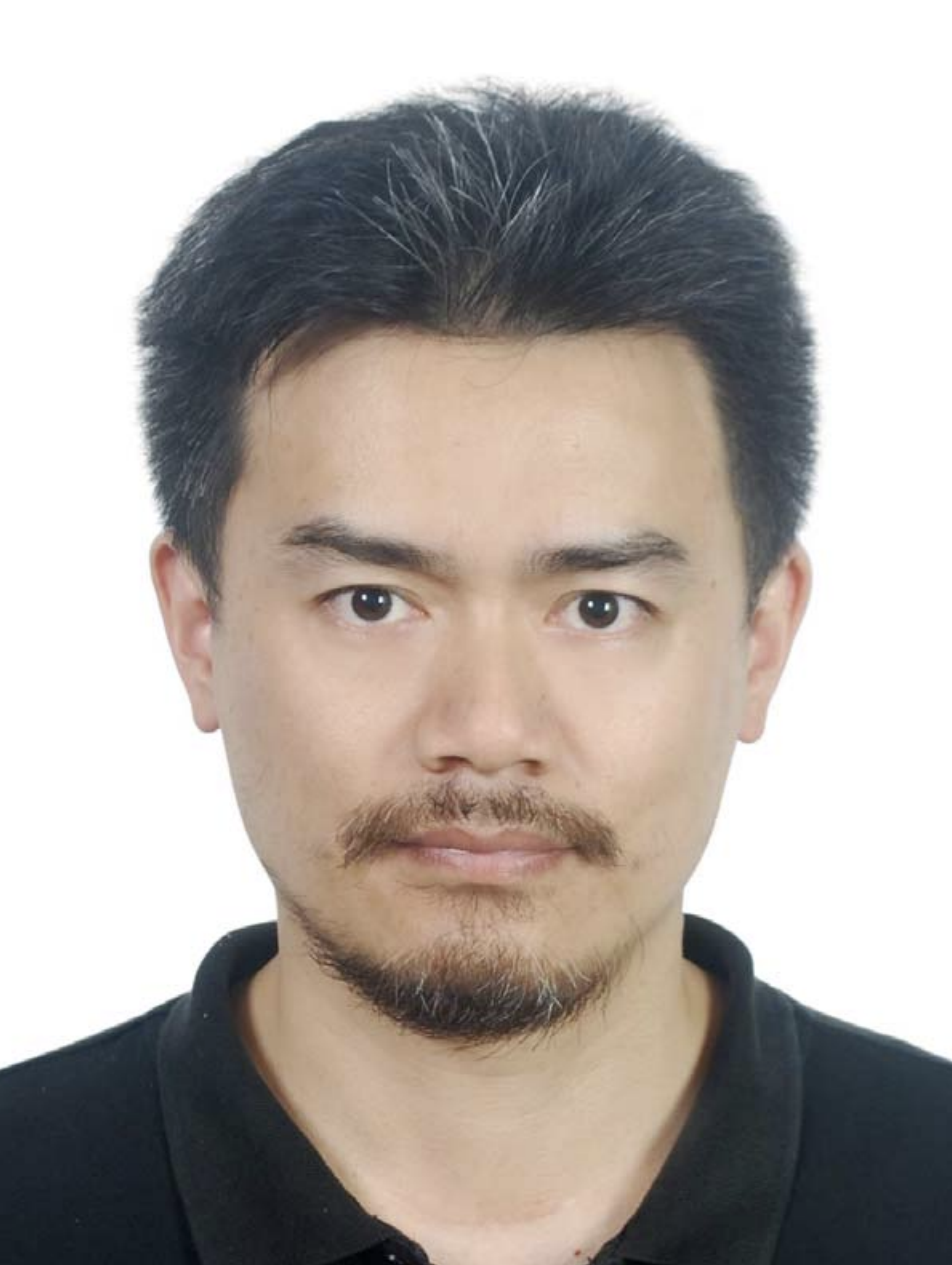}}
\end{wrapfigure}\par
\noindent \textbf{Kai Huang} received the B.Sc. degree from Fudan University in 1999, the M.Sc. degree from University Leiden in 2005, and the Ph.D. degree from ETH Zurich in 2010. He joined Sun Yat-Sen University, Guangzhou, China as a Professor in 2015. He was a Senior Researcher with the Computer Science Department, Technical University of Munich, Munich, Germany from 2012 to 2015, and a Research Group Leader with Fortiss GmbH, Munich, Germany, in 2011. His research interests include techniques for the analysis, design, and optimization of embedded/CPS systems, particularly in the automotive, medical, and robotic domains. Prof. Huang was a recipient of best paper awards/candidates for a number of conferences.\par

\end{document}